\documentclass[lettersize,journal]{IEEEtran}
\usepackage{amsmath,amsfonts}
\usepackage{algorithmic}
\usepackage{algorithm}
\usepackage{array}
\usepackage[caption=false,font=normalsize,labelfont=sf,textfont=sf]{subfig}
\usepackage{textcomp}
\usepackage{stfloats}
\usepackage{url}
\usepackage{verbatim}
\usepackage{graphicx}
\usepackage{cite}
\usepackage{algorithm}
\usepackage{algorithmic}
\usepackage{epsfig}
\usepackage{amsmath}
\usepackage{amssymb}
\usepackage{xcolor}
\usepackage{multirow}
\usepackage{adjustbox}
\usepackage{booktabs}
\usepackage{soul}
\definecolor{lightyellow}{rgb}{0.95,0.95,0.5}
\sethlcolor{lightyellow}
\usepackage{multirow}
\usepackage{comment}

\newcommand{\methodname}{Class-Adaptive Cross-Attention}
\newcommand{\methodacro}{\emph{(CA)$^2$-SIS}}

\begin{document}

\title{Semantic Image Synthesis via \\ Class-adaptive Cross-attention}

\author{Tomaso Fontanini, Claudio Ferrari, Giuseppe Lisanti, Massimo Bertozzi, Andrea Prati
\thanks{This paper was produced by the IEEE Publication Technology Group. They are in Piscataway, NJ.}
\thanks{Manuscript received April 19, 2021; revised August 16, 2021.}}

\markboth{Journal of \LaTeX\ Class Files,~Vol.~??, No.~?, October~2023}%
{Shell \MakeLowercase{\textit{et al.}}: A Sample Article Using IEEEtran.cls for IEEE Journals}


\maketitle

\begin{abstract}
In semantic image synthesis the state of the art is dominated by methods that use customized variants of the SPatially-Adaptive DE-normalization (SPADE) layers, which allow for good visual generation quality and editing versatility. By design, such layers learn pixel-wise modulation parameters to de-normalize the generator activations based on the semantic class each pixel belongs to. Thus, they tend to overlook global image statistics, ultimately leading to unconvincing local style editing and causing global inconsistencies such as color or illumination distribution shifts. Also, SPADE layers require the semantic segmentation mask for mapping styles in the generator, preventing shape manipulations without manual intervention. In response, we designed a novel architecture where cross-attention layers are used in place of SPADE for learning shape-style correlations and so conditioning the image generation process. Our model inherits the versatility of SPADE, at the same time obtaining state-of-the-art generation quality, as well as improved global and local style transfer. Code and models available at https://github.com/TFonta/CA2SIS.
\end{abstract}

\section{Introduction}\label{sec:introduction}
Semantic image synthesis is the task of generating realistic images conditioned on a semantic mask, \textit{i.e.} a pixel-wise annotation of the semantic classes defining the spatial layout. Since the initial stages of exploration in this domain, the predominant trend consisted in using spatially-adaptive normalization layers, firstly proposed by Park~\textit{et al.} and known as SPADE~\cite{park2019semantic}. Those are designed to modulate the activations in the generator layers to propagate the semantic information, and so condition the generated samples in a spatially-adaptive manner. This mechanism proved effective in terms of generation quality and semantic control.
Under this paradigm, images can be generated either by encoding the style, \textit{i.e.} texture, from a reference image (\textit{reference-based}), or in a fully generative setting (\textit{diversity-based}). Most methods belong to the latter set, where the objective is to produce realistic yet diverse outputs given the same semantic mask~\cite{richardson2021encoding,wang2021image,park2019semantic,tan2021diverse}. Whereas reference images are usually employed during training and can be used to guide the generation, such methods focus primarily on multi-modality and diversity. Oppositely, reference-based approaches tackle the problem of encoding the style of a specific image, with the primary goal of accurately reconstructing and editing \textit{real images}~\cite{lee2020maskgan, zhu2020sean}. The two settings, despite similar in their goals and often sharing architectural design, demand for dedicated solutions to deal with different challenges. In diversity-driven approaches, some essential objectives are being able to generate multi-modal outputs given the same semantic layout, or avoiding overfitting. It is quite common that, for example, if a semantic mask of a human face has long hair, then the model will output an image of a woman, even if that information is completely unknown. On the opposite, reference-based methods deal with totally different issues. For instance, a good reference-based method should be able to retain, in the case of human faces, the perceived identity of the portrayed individual. In this paper, we are specifically interested in the reference-based scenario. Most parts of the paper will hence be focused on this aspect, yet examples in the diversity-driven setting will be provided as well.

Despite alternatives have been explored in the literature, such as using layout-to-image conditional convolutions~\cite{wang2021image,liu2019learning}, SPADE and its variants still represent the standard choice in the field of semantic image synthesis. Even very recent approaches such as Semantic Diffusion~\cite{wang2022semantic} employ SPADE layers to condition the generation with the semantic layout. However, a drawback of all SPADE-based approaches (both diversity- and reference-based) is the tendency to introduce overall inconsistencies in the generated images. Indeed, they apply the feature modulation through learning class-wise normalization parameters, independently for each semantic class. Thus, global statistics such as illumination or color distribution, as well as long-range dependencies, are neglected, ultimately leading to inconsistencies in the generated images.


To address the above, in this paper we propose a significant paradigm change for semantic image synthesis, to answer the question on how to simultaneously improve the generation quality while maintaining a fine-grained control over the semantic classes. Specifically, we explore the use of cross-attention~\cite{vaswani2017attention,chen2021crossvit} as an alternative to SPADE for conditioning the image generation in a generative adversarial setting. Cross-attention layers proved mostly effective when used with diffusion models to condition the image generation via text embeddings~\cite{saharia2022photorealistic, rombach2022high}. They allow the conditioning mechanism to be: \textit{(a)} very flexible, since cross-attention lets any intermediate representation to be mapped inside the network; \textit{(b)} much more consistent, since attention accounts for long-range dependencies in the input data. Our proposed model naturally blends the capability of class-level style control as in previous GAN methods based on SPADE, with the versatility, improved quality and consistency of cross-attention.  

In our framework, the input to the generator is the semantic mask, while the style features extracted from a reference image are the condition to the cross-attention layers. 
In SPADE layers, the spatial layout of the semantic mask can be directly used to apply class-specific feature normalization at precise spatial locations defined by the mask pixels, which forces a strong spatial consistency in the output image. On the other hand, it represents a noticeable constraint as inter-class dependencies or global statistics cannot be captured. 
Cross-attention layers allow instead to learn shape-style correlations by injecting style information into the model while the semantic mask can be encoded into a latent code and used as input to the whole system. We will show that this is advantageous in many ways: it leads to increased robustness to inaccuracies in the semantic masks, while also improving the overall image consistency. Being the mask encoded into a latent feature, it also enables to perform latent manipulation so to edit \textit{the shape} other than the style. Further collateral advantage of this solution is the prospect of conditioning the generator with arbitrary style features. This versatility allowed us to design a specific style encoder that extracts multi-scale features and is equipped with class-adaptive grouped convolutions to optimize the representation of each class. In sum, the main contributions of this work are:
\begin{itemize}
    \item we explore the use of cross-attention in place of SPADE layers in reference-based semantic image synthesis, and design a novel architecture to blend such mechanism into a GAN framework, inheriting the advantages of both. To this aim, we also implemented a novel multi-resolution style encoder equipped with group convolutions;
    \item we introduce a novel training mechanism with cross-attention by employing an attention loss that forces the learned attention maps to match the shape of the semantic mask, increasing the style controllability; 
    \item we demonstrate the advantage of cross-attention versatility by designing a new style encoder optimized for extracting multi-scale, class-level style features that can be readily plugged into the generator architecture;
    \item we extensively show that the proposed solution improves upon the state of the art, and provides an alternative paradigm that brings several advantages over prior works.
\end{itemize}

\section{Related Work}\label{sec:related}
From a technical standpoint, all semantic image synthesis methods in the literature share similar frameworks. The two most common modules are the Style Encoder and the Generator network. The former is responsible for encoding the style (either from a reference image or noise), while the generator progressively up-samples the semantic layout provided as input. The output image is generated by injecting the encoded style at specific spatial locations defined by the semantic mask. 
Notwithstanding the different architectural choices, the major challenge is to properly encode and inject the style information into the generator.  In this regard, Park~\textit{et al.}~\cite{park2019semantic} first noted that in conventional synthesis architectures~\cite{isola2017pix2pix,wang2018cgan}, the normalization layers tend to remove the information contained in the input semantic masks. They thus proposed the SPatially-Adaptive (DE)normalization (SPADE) method to overcome this problem. The same model, also known as GauGAN~\cite{park2019gaugan}, was deployed in a GAN-based image synthesis application.
Since the introduction of SPADE, a lot of effort has been put into investigating its limitations and finding solutions for improvement.
For example, Zhue~\textit{et al.}~\cite{zhu2020sean} noted that in SPADE only a single style code is used to control the style of the whole image. To gain more fine-grained generation control, they designed a SEmantic region-Adaptive Normalization block (SEAN), allowing to control the style of each semantic class individually.
Simultaneously, Lee~\textit{et al.}~\cite{lee2020maskgan} proposed a similar framework specific for human faces, named MaskGAN, to enable diverse and interactive face manipulation. Nevertheless, SEAN outperforms MaskGAN in almost every aspect.  
Similar to SEAN, Tan~\textit{et al.}~\cite{tan2021efficient} proposed a CLass-Adaptive (DE)normalization layer (CLADE) that uses the input semantic mask to modulate the normalized activation in a class-adaptive manner. 
Later, to further push the style control beyond the class level, the same authors proposed the INstance-Adaptive DEnormalization (INADE)~\cite{tan2021diverse} approach that is capable of producing diverse results even at the instance level. Both CLADE and INADE are also equipped with an extra style encoder trained with KL loss to provide quality-driven results (V-CLADE, V-INADE). Along this line, several other works were proposed~\cite{liu2019learning, fontanini2023automatic, fontanini2023frankenmask, li2021collaging, shi2022retrieval}. Finally, even recent methods based on diffusion models were proposed that use SPADE layers to achieve the conditioning process. An example is Semantic Diffusion~\cite{wang2022semantic}, which adapted the diffusion process to be conditioned with semantic layouts.
From a different perspective, Sushko \textit{et al.}~\cite{sushko2022oasis} proposed a SPADE-based architecture where an alternative training paradigm was designed that relies solely on adversarial supervision.

All the above mentioned papers employ SPADE or its variants. Other works attempted to define alternative solutions such as SC-GAN~\cite{wang2021image}, which introduced a novel semantic encoding and stylization methods via spatially-variant and appearance-correlated operations, inspired by the layout-to-image conditional convolutions earlier proposed by Liu \textit{et al.}~\cite{liu2019learning}.
Other alternatives take inspiration from StyleGAN~\cite{karras2019style} and adapt it for the semantic synthesis task. Specifically, we mention SemanticStyleGAN~\cite{shi2021semanticstyle}, which is a StyleGAN-based architecture that models local semantic parts separately via learning structural priors through the semantic mask.

Differently from all the above, in this work we revisit the way in which style is injected into the generator by exploring the use of self- and cross-attention in place of SPADE, taking inspiration from recent latent diffusion models~\cite{rombach2022high}. The latter use external inputs, \textit{e.g.} text, audio, or semantic masks, to condition the diffusion process, whereas we propose an opposite paradigm where encoded \textit{style features} are the condition, and the semantic layout is the input to the generative process. 
In doing so, we take advantage of the versatility of such layers while simultaneously avoiding the extremely-large computational cost that is required to train diffusion models. Moreover, our solutions enables local style control over the image generation, which is quite hard to obtain with current diffusion models, even if conditioned with semantic masks as in Semantic Diffusion~\cite{wang2022semantic}. 

Among all the reference works, one method that stands out in terms of similarity with ours is SEAN~\cite{tan2021efficient}, which is the only one purely reference-based. Despite most of the other methods can use some reference image to sample the style features, they are mostly diversity-based. Thus, in the remainder of this paper, we will mostly refer to SEAN as representative of SPADE-based methods for comparison.


\section{Architecture}
\label{sec:architecture}
\begin{figure*}[t]
    \centering
    \includegraphics[width=0.8\linewidth]{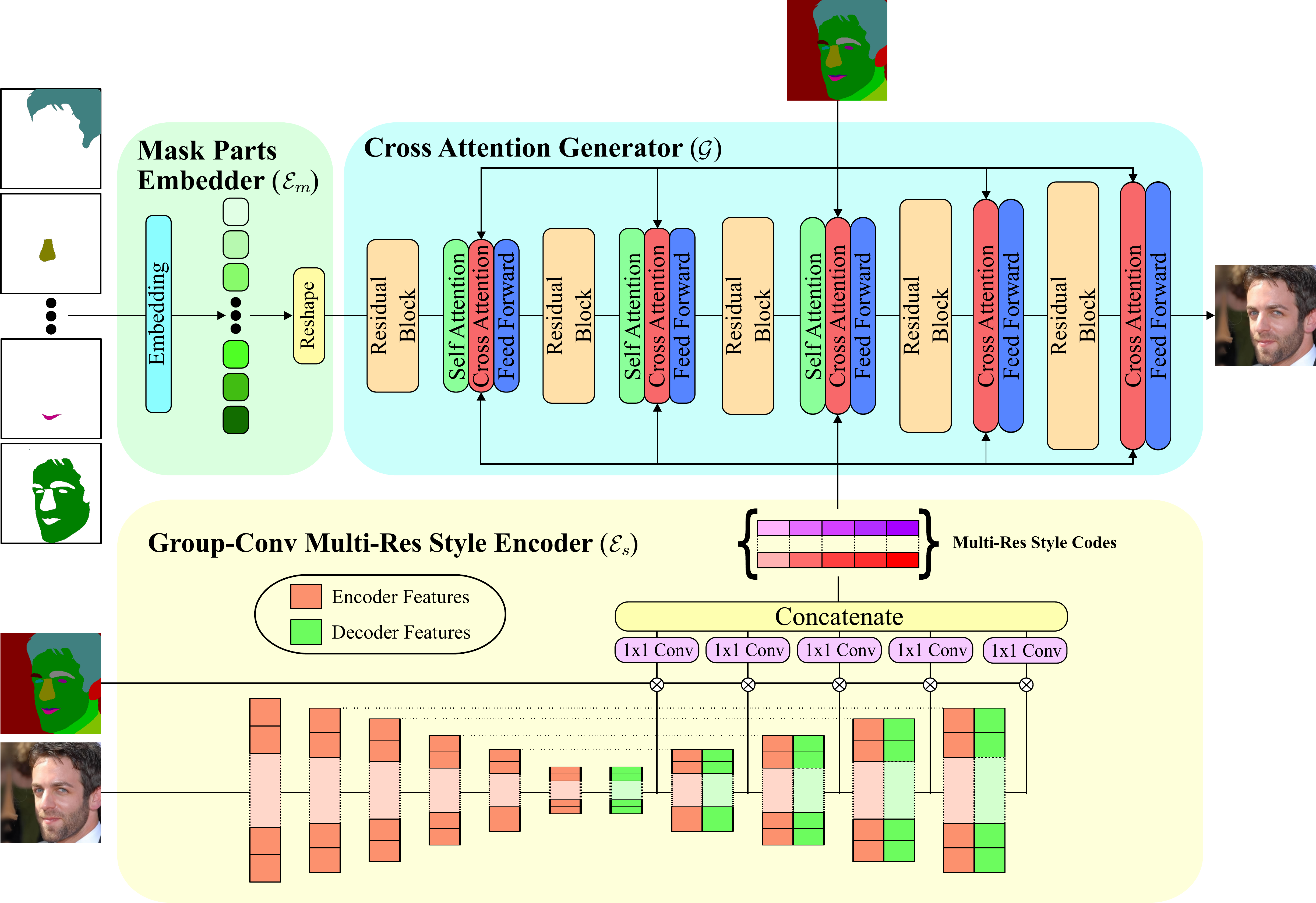}
    \caption{\methodacro{} architecture: style codes are extracted using a Multi-Resolution Style Encoder $\mathcal{E}_s$ equipped with grouped convolutions; the Mask Embedder $\mathcal{E}_m$ embeds each of the semantic mask parts into a set of latent codes; finally, these codes are fed to the Cross-Attention Generator $\mathcal{G}$ that is conditioned with the style codes thanks to the cross-attention mechanism. Additionally, the semantic mask is also used in the cross-attention layer to calculate the attention loss $\mathcal{L}_{att}$ in order to push each attention map to follow the mask shape. }
    \label{fig:arch}
\end{figure*}

The proposed architecture (see Fig.~\ref{fig:arch}) is called \methodname{}, \methodacro{}, and is composed by three main modules: (a) a Multi-Resolution Style Encoder $\mathcal{E}_s$, with Grouped Convolutions, Group Normalization layers and skip connections, that is used to extract style features from an RGB reference, (b) a Mask Embedder $\mathcal{E}_m$ that extracts a latent representation separately for each semantic class and, finally, (c) a Cross-Attention Generator $\mathcal{G}$ that exploits the attention mechanism to inject the multi-resolution style codes inside the network in order to control the generated image. Additionally, a Discriminator $\mathcal{D}$ is employed during training to enforce the adversarial loss.

\subsection{Multi-Resolution Grouped Style Encoder}
\label{subsec:style-encoder}
The purpose of the style encoder is to extract style features from an RGB image $y \in \mathbb{R}^{H\times W \times 3}$ to condition the generator. 
In order to independently control the style of different semantic classes, we improve the SEAN~\cite{zhu2020sean} style encoder, which uses the semantic mask to adaptively pick style features from specific areas of the RGB image. 
However, SEAN and other methods based on SPADE~\cite{park2019semantic} extract style features only from the last layer of the encoder. We observed that this led to inconsistent style mapping for classes at different scales. This motivated us to consider a different design; in particular, we defined a deeper encoder to extract features at multiple scales. Although using multi-scale features was explored for mapping style features to a pre-trained StyleGAN generator~\cite{richardson2021encoding}, here we build further improvements upon the versatility brought by cross-attention. Specifically, we enhance the feature representation of each class by also employing grouped convolutions. 

\noindent
\textbf{Grouped convolutions}: Let a semantic mask be a $C$-channel image $\mathcal{M} \in \mathbb{N}^{C \times H \times W}$, where each channel $\mathcal{M}_j$ is a binary image encoding the spatial location of a specific class. 
Then, we define the $i$-th convolutional layer of the style encoder to have $(C \times f)$ filters \textit{i.e.} one group for each semantic class, each group having $f$ filters with their own learnable weights. 
Differently from previous solutions~\cite{zhu2020sean} which process all features together for each layer with classic convolutions, this design allows us to sample styles from different feature groups, each relative to a specific semantic class. 
Each layer is followed by group normalization, which learns specific statistics related to the features of a specific mask channel, and by a ReLU activation function. 

\noindent
\textbf{Multi-resolution feature pooling}: Let the style features of group $j$ resulting from the $i$-th layer of the encoder be $\mathcal{F}_{i,j} \in \mathbb{R}^{f_j \times H_i \times W_i}$, then the class-wise style codes are extracted with an average pooling (AP) of the masked features:
\begin{equation}
    S_{i, j} = AP\left(\mathcal{F}_{i,j} \cdot \mathcal{M}_j\right)
\end{equation}

\noindent
Note that, depending on the resolution of the $i$-th layer, the mask $\mathcal{M}$ is resized accordingly. 
This is done for each upsampling layer $i = 1, \cdots, L_{up}$ of the encoder and for each semantic class \textit{i.e.} mask channel, $j$. 
Pooled features $S_{i, j}$ are then reshaped using $1 \times 1$ convolutions, resulting in codes of dimension 256, and then concatenated to form a size of $C \times (L_{up} \cdot 256)$. 

The overall architecture of our grouped multi-resolution style encoder ($\mathcal{E}_s$) is composed of $6$ down-sampling layers and $5$ up-sampling layers (i.e., $L_{up}$), linked together using skip connections. 
So, the final style codes have a dimension of $C \times 1280$. 

\subsection{Mask Parts Embedder}\label{subsec:mask-embedder}
All SPADE-based semantic image synthesis methods use the raw semantic masks as input to the de-normalization layers since they require spatial information to apply feature maps modulations.
With cross-attentions we get rid of this constraint. We hence pass the semantic mask through a Mask Embedder ($\mathcal{E}_m$) to obtain a latent representation. We will show (Table~\ref{tab:shape_transfer}, Fig.~\ref{fig:swap_parts}) that this leads to both quantitative and qualitative improvements.
The design of $\mathcal{E}_m$ is very simple yet effective: we flatten each mask channel, and pass them \textit{separately} through a linear layer, producing a latent code $m_j$ of dimension $256$ for each of the $C$ channels. 
An MLP network was chosen instead of a convolutional one as MLPs were shown invariant to pixel-level transformations such as translation or shuffling~\cite{ivan2019convolutional}. These $C$ codes are then reshaped to form a $C \times 16 \times 16$ representation of the mask, that can be finally used as input to the generator.

\begin{figure}
    \centering
    \includegraphics[width=\linewidth]{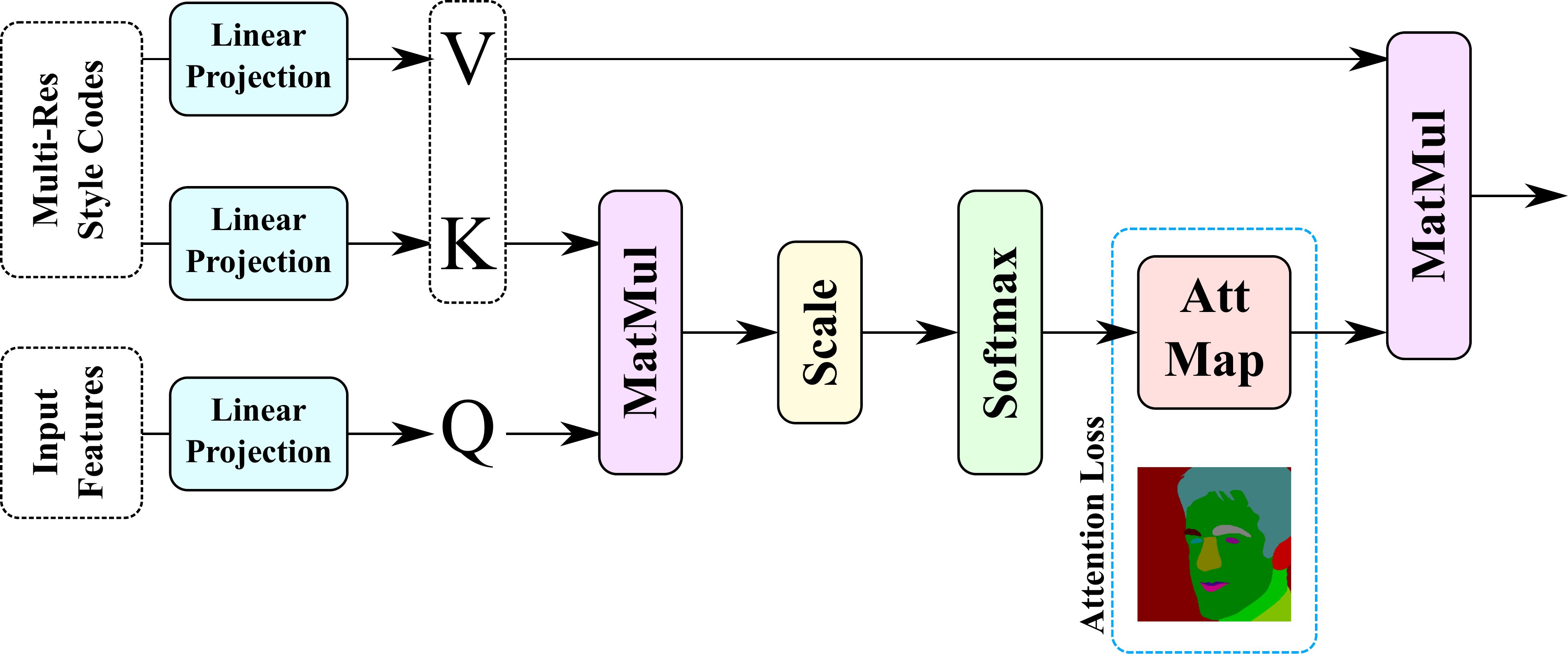}
    \caption{Cross-Attention layer: The Query $Q$ is derived from the features of the previous residual block, while Key $K$ and Value $V$ are calculated starting from the multi-resolution style codes. Additionaly, an attention loss between the output of the Softmax and the semantic map $\mathcal{L}_{att}$ is calculated.}
    \label{fig:ca}
\end{figure}

\subsection{Cross Attention Generator}\label{subsec:generator}

The generator network is the main contribution of this work, as it significantly differentiates from the previous literature where custom normalization layers were used. Our generator is made up of 5 up-sampling blocks, each composed by a convolutional residual block and a spatial transformer block. More in detail, the spatial transformer blocks include a self-attention layer, a cross-attention layer and a feed-forward layer, following the structure of~\cite{rombach2022high}. Given that the memory footprint of self-attention layers is quadratic with respect to the input, we use self-attention up to feature maps of size $64\times 64$, therefore the last two attention blocks only employ cross-attention. Attention is defined as follows:
\begin{equation}
     \mathcal{A}(Q,K,V) = \mathcal{M}_A(Q,K) \cdot V
     \label{eq:attention}
\end{equation}
\noindent
where $\mathcal{M}_A(Q,K)$ is the attention map, defined as:
\begin{equation}
    \mathcal{M}_A(Q,K) = \mathcal{S}\left(\frac{QK^T}{\sqrt{d}}\right)
    \label{eq:att_map}
\end{equation}
\noindent
$d$ is the output dimension of each attention head in a multi-head configuration ($d=64$, as in~\cite{vaswani2017attention}), and $\mathcal{S}()$ is the Softmax activation function. In self-attention, $Q, K$ and $V$ are all obtained from the projection of the same embeddings, that is the flattened features $\phi^{(i)}$ of the generator at layer $i$:
\begin{equation}
    Q =  W_Q^{(i)} \cdot \phi^{(i)}, 
    K =  W_K^{(i)} \cdot \phi^{(i)}, 
    V =  W_V^{(i)} \cdot \phi^{(i)}
\end{equation}
\noindent
 Cross-attentions are meant to map the styles extracted by the encoder into the generator, in order to condition the generated samples. Hence, in cross-attention, $K,V$ are computed from the style codes $\mathcal{E}_s(y)$ as follows:
\begin{equation}
    Q=W_Q^{(i)} \cdot \phi^{(i)}, 
    K = W_K^{(i)} \cdot \mathcal{E}_s(y), 
    V = W_V^{(i)} \cdot \mathcal{E}_s(y)
\end{equation}

\smallskip
\noindent\textbf{Attention Loss.} By mixing style features $\mathcal{E}_s(y)$ with the flattened generator features $\phi^{(i)}$ which encode layout information from the semantic mask, cross-attentions implicitly learn how to map the style of each class into the relative spatial locations. This mapping is learned and encoded in the attention map $\mathcal{M}_A(Q,K)$. However, both self- and cross-attention compute long-range global dependencies, meaning that they capture correlations across different classes. This results in the attention maps $\mathcal{M}_A(Q,K)$ to spread beyond the class boundaries defined by the segmentation mask $\mathcal{M}$. In fact, certain styles might be class-wise correlated, \textit{e.g.} hair color and skin tone, which is captured by the attention maps. On the one hand, this helps preserving the overall image quality and realism by forcing consistency across styles; on the other, it limits local style controllability (see Fig.~\ref{fig:swap_att}). 


In order to maximize both quality and controllability, during training we impose a loss term to push the learned attention maps $\mathcal{M}_A(Q,K)$ to follow the shape of the semantic mask $\mathcal{M}$ (Fig. \ref{fig:ca}). This allows to increase the style controllability without sacrificing the generation quality. The proposed \emph{attention loss} is a binary cross-entropy (BCE) computed between the attention map and the semantic mask. Since, in cross-attention, the key values are the projected style codes, the attention map $\mathcal{M}_A(Q,K)$ has size $h \times C \times H \times W$, where $h$ is the number of attention heads, $C$ is the number of semantic classes and $H,W$ the height and width of the feature maps. Thus, it is possible to impose that all the $h$ attention maps of the $c$-th semantic class match the $c$-th input mask channel pixel-wise. The loss for a single semantic class $c$ is computed as:
\begin{equation}
 \label{eq:att_loss}
 \resizebox{\linewidth}{!}{%
 $
 \begin{split}
    \mathcal{L}_{att}^c = \frac{1}{HWh}\sum_{i=1}^{H}\sum_{j=1}^{W}\sum_{k=1}^{h} y_{cij}^k\log(x_{cij}^k) 
    + (1-y_{cij}^k)\log(1-x_{cij}^k)
\end{split}
$
}
\end{equation}

\noindent
where $y_{cij}^k$ is the class of pixels at channel $c$, row $i$, column $j$, for the $k$-th head, and $x_{cij}^k$ the corresponding attention map value. Fig. \ref{fig:swap_att} shows how, by imposing this loss when swapping styles, the entanglement is greatly reduced. 

\begin{table*}[]
    \centering
    \begin{adjustbox}{width=0.8\textwidth,center}
    \begin{tabular}{c|c|c|c|c|c|c|c|c|c@{}}
    \toprule
    & \multicolumn{3}{c|}{CelebMask-HQ} & \multicolumn{3}{c|}{Ade20k} & \multicolumn{3}{c}{DeepFashion}\\
    \hline
    \textbf{Method} & \textbf{FID} $\downarrow$ & \textbf{mIoU} $\uparrow$ & \textbf{Acc.} $\uparrow$ & \textbf{FID} $\downarrow$ & \textbf{mIoU} $\uparrow$ & \textbf{Acc.} $\uparrow$  & \textbf{FID} $\downarrow$ & \textbf{mIoU} $\uparrow$ & \textbf{Acc.} $\uparrow$\\
    \hline 
    Pix2PixHD~\cite{wang2018high} & 22.26 & 78.40 & 92.88 & 66.65 & 28.47 & 63.78 & 15.33 & 89.52  & 98.47 \\
    SPADE~\cite{park2019semantic} & 21.08 & 78.32 & 92.76 & 53.70 & 44.21 & 69.05 & 11.18 & \hl{\textbf{92.87}}  & \hl{\textbf{99.11}} \\
    MaskGAN~\cite{lee2020maskgan} & 59.91 & 76.34 & 87.89 & * & * & * & * & * & * \\  
    SEAN~\cite{zhu2020sean} & 18.72 & \hl{\textbf{78.62}} & \hl{93.54} & \hl{38.63} & 43.82 & 67.42 & 
    10.70 & 92.19 & 98.72 \\
    V-INADE~\cite{tan2021diverse} & \hl{17.49} & 78.04 & 93.50 & 39.87 & \hl{45.93} & \hl{\textbf{70.58}} & \hl{10.33} & 92.40 & 98.85\\
    \hline    
    \textbf{Ours} - w/o AttLoss & \hl{\textbf{15.80}} & 78.01 & 93.42 & \hl{\textbf{38.08}} & 45.72 & 69.04 & \hl{\textbf{9.97}} & 90.95 & 98.56 \\
    \textbf{Ours} - Full & 15.84 & \hl{78.54} & \hl{\textbf{93.78}} & 38.34 & \hl{\textbf{46.05}} & \hl{69.27} & 10.63 & \hl{91.24} & \hl{98.61} \\
    \bottomrule   
    \end{tabular}
    \end{adjustbox}
    \caption{Comparison with the state-of-the-art in terms of FID, mIoU and segmentation accuracy (\textbf{Acc.}). 
    Best results in bold. \hl{Highlight} indicates the best result among our variants and the competitors. ``*'' indicates MaskGan is trainable only for CelebMask-HQ.}
    \label{tab:celeba-rec}
\end{table*}

\section{Experimental Results}\label{sec:experiments}

\noindent\textbf{Implementation details}. 
The objective of our model is to reconstruct a target RGB image from its semantic mask.
This is enforced by means of an adversarial loss, a feature matching loss~\cite{wang2018cgan} and a perceptual loss~\cite{johnson2016perceptual}. In addition, an attention loss is proposed and added as described in Sec. \ref{subsec:generator}.
We employ a multi-scale discriminator~\cite{park2019semantic} and train the model for 100 epochs on a NVIDIA A100 GPU, using the Adam optimizer with a learning rate of 0.0002. Generated images have size $256 \times 256$.

\noindent\textbf{Datasets}. 
Experiments are conducted on three public datasets: CelebAMask-HQ~\cite{lee2020maskgan}, Ade20k~\cite{zhou2017scene} and DeepFashion~\cite{liu2016deepfashion}.

\begin{figure}[!t]
    \centering
    \includegraphics[width=\linewidth]{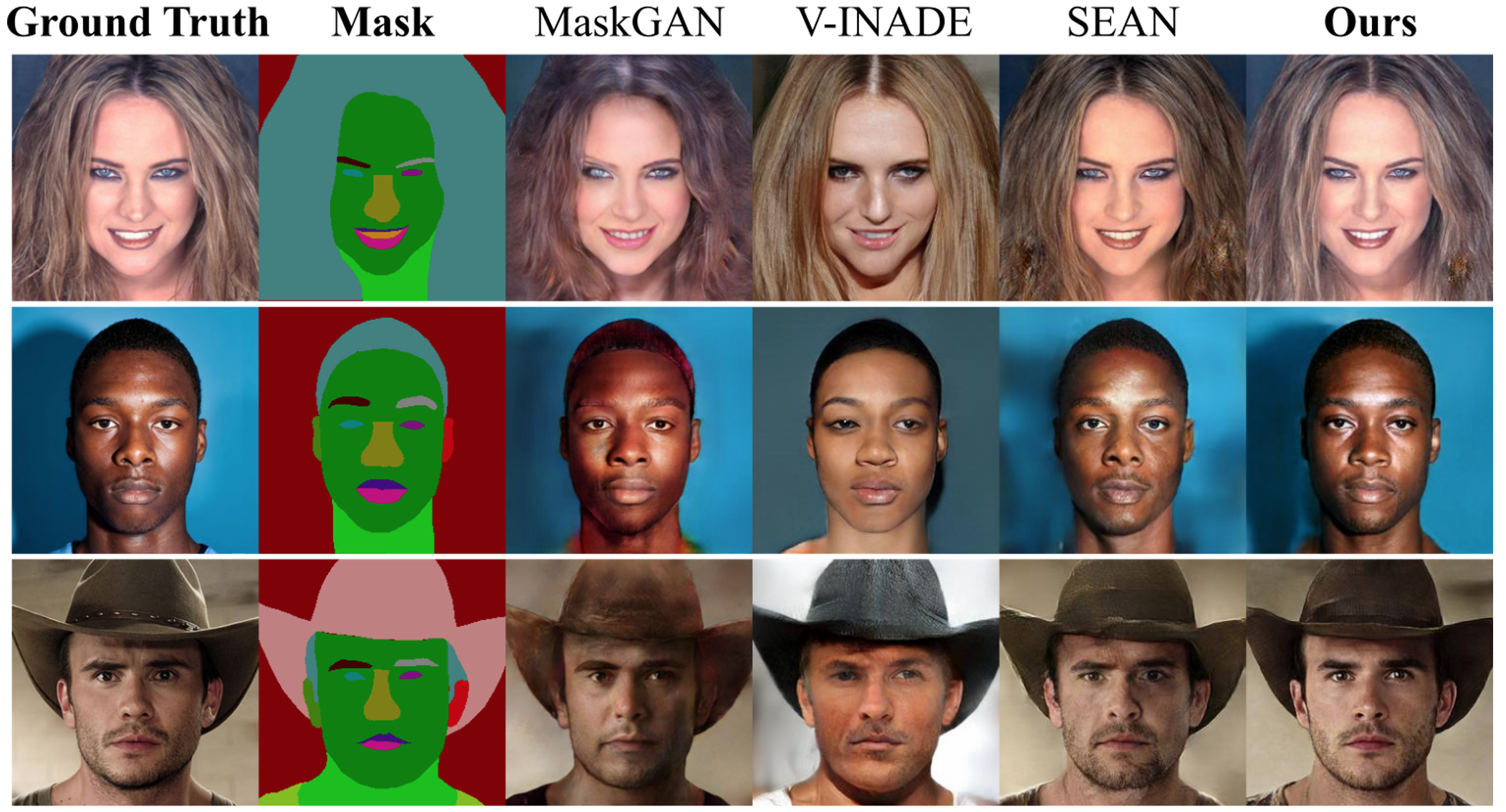}
    \caption{Qualitative comparison between state-of-the-art methods and our architecture on CelebMask-HQ. Our approach better preserves the color distribution (top row) and illumination coherence (second row).}
    \label{fig:rec}
\end{figure}

\noindent\textbf{Metrics}. Following previous works~\cite{zhu2020sean,lee2020maskgan,park2019semantic}, we employ the following metrics for evaluating our method. First, we use the Frechet Inception Distance (FID), to estimate the generation quality. 
Additionally, a semantic-segmentation-based evaluation was used to compare the ground-truth layouts against those obtained by running a pre-trained segmentation method on the generated images, in terms of mean Intersection-over-Union (mIoU) and pixel accuracy. 
Following previous works, we employ \emph{FaceParsing}\footnote{{\scriptsize https://github.com/switchablenorms/CelebAMask-HQ/tree/master/face\_parsing}} for CelebAMask-HQ and DeepFashion, and \emph{SceneSegmetantion}\footnote{{\scriptsize https://github.com/CSAILVision/semantic-segmentation-pytorch}} for Ade20k, respectively. 

\subsection{Reconstruction}
We first report a quantitative comparison against state-of-the-art works in terms of semantic synthesis quality (Table~\ref{tab:celeba-rec}) of two versions of the proposed method: with and without attention loss. As discussed in Sect.~\ref{sec:related}, most methods are diversity-driven, with the exception of SEAN~\cite{zhu2020sean} which is purely reference-based. However, some of them are equipped with a style encoder that allows them to use a reference image. Thus, here we provide a comparison with such, in particular 
Pix2PixHD~\cite{wang2018high}, SPADE~\cite{park2019semantic}, MaskGAN~\cite{lee2020maskgan}, SEAN~\cite{zhu2020sean} and INADE~\cite{tan2021diverse} when equipped with its style Variational autoencoder (V-INADE). A comparison against purely diversity-driven approaches, \textit{i.e.}~SC-GAN~\cite{wang2021image}, INADE~\cite{tan2021diverse}, and SemanticDiffusion~\cite{wang2022semantic} will be presented in Sect.~\ref{subsec:diversity}. 

\begin{figure}[!t]
    \centering
    \includegraphics[width=\linewidth]{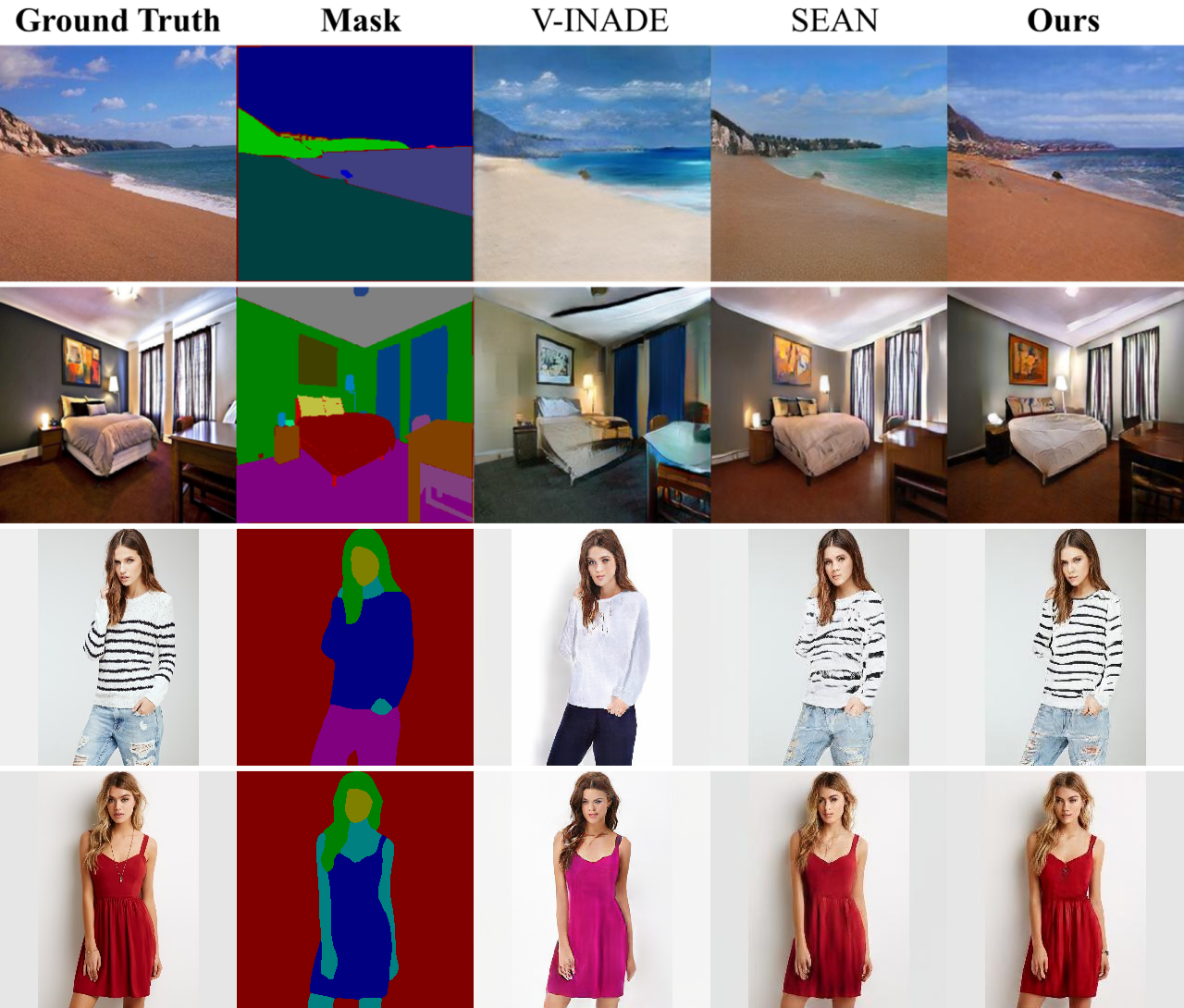}
    \caption{Qualitative comparison between state-of-the-art methods and our architecture on Ade20k and DeepFashion. Our approach better preserves the color distribution (top row) and illumination coherence (second and bottom rows). MaskGAN is not shown since is trainable only for face images.}
    \label{fig:ade}
\end{figure}

Table~\ref{tab:celeba-rec} shows that our novel solution based on shape-style attention blocks performs competitively with respect to state-of-the-art methods, obtaining the best FID score in all datasets. Interestingly, when adding the attention loss, FID score gets slightly worse; this can be explained by the fact that imposing stronger disentanglement in the attention maps means less freedom in the generation, which is always reflected in a slightly degraded FID score. Unfortunately though, the FID score does not fully reflect the quality improvement obtained by using cross-attentions in place of SPADE, which can be better appreciated in the qualitative examples in Fig.~\ref{fig:rec} and~\ref{fig:ade}. The reader can appreciate how our method better preserves the overall color distribution such as the skin tone (Fig.~\ref{fig:rec} top row) or sand color (Fig.~\ref{fig:ade} top row), and more complex details such as the cast shadow  (Fig.~\ref{fig:rec}, second row), or the clothing folds (Fig.~\ref{fig:ade}, bottom row). To better highlight this point, more examples can be found in the supplementary material.

On the other hand, we obtain slightly lower performance for segmentation accuracy measures (mIoU and accuracy). A reasonable explanation is that SPADE provides hard pixel-wise guidance by using the semantic mask (resized according to the size of each layer) in \textit{all} the generator layers to modulate the activations at specific spatial locations. This is done both during training and inference. Instead, we \textit{do not directly inject} layout information in the generator, but only use the (embedded) semantic mask as input, and during training for computing the attention loss $\mathcal{L}_{att}$, providing weaker supervision. Nonetheless, performance are totally comparable. This is a clear hint that, even without the need of an explicit spatial mapping, cross-attention do automatically learn meaningful shape-style relationships. This downside is partially avoided when employing the attention loss; imposing a constraint to localize the attention maps has a positive effect over the shape related measures.

Finally, it is worth noting that for CelebAMask-HQ our method performs significantly better compared to all methods in terms of FID score. One potential explanation is that in CelebAMask-HQ there is a higher level of style correlation compared to other datasets. In human faces, indeed, eyes, hair and skin tone are highly correlated. Similarly, older people are more likely to have grey/white hair, and this correlation is captured by attention blocks. The same applies to a lesser extent for Ade20k, mostly for outdoor scenes, where grass color is likely correlated with that of the trees, or sea color with that of the sand. 

\begin{figure}[t]
    \centering
    \includegraphics[width=.99\linewidth]{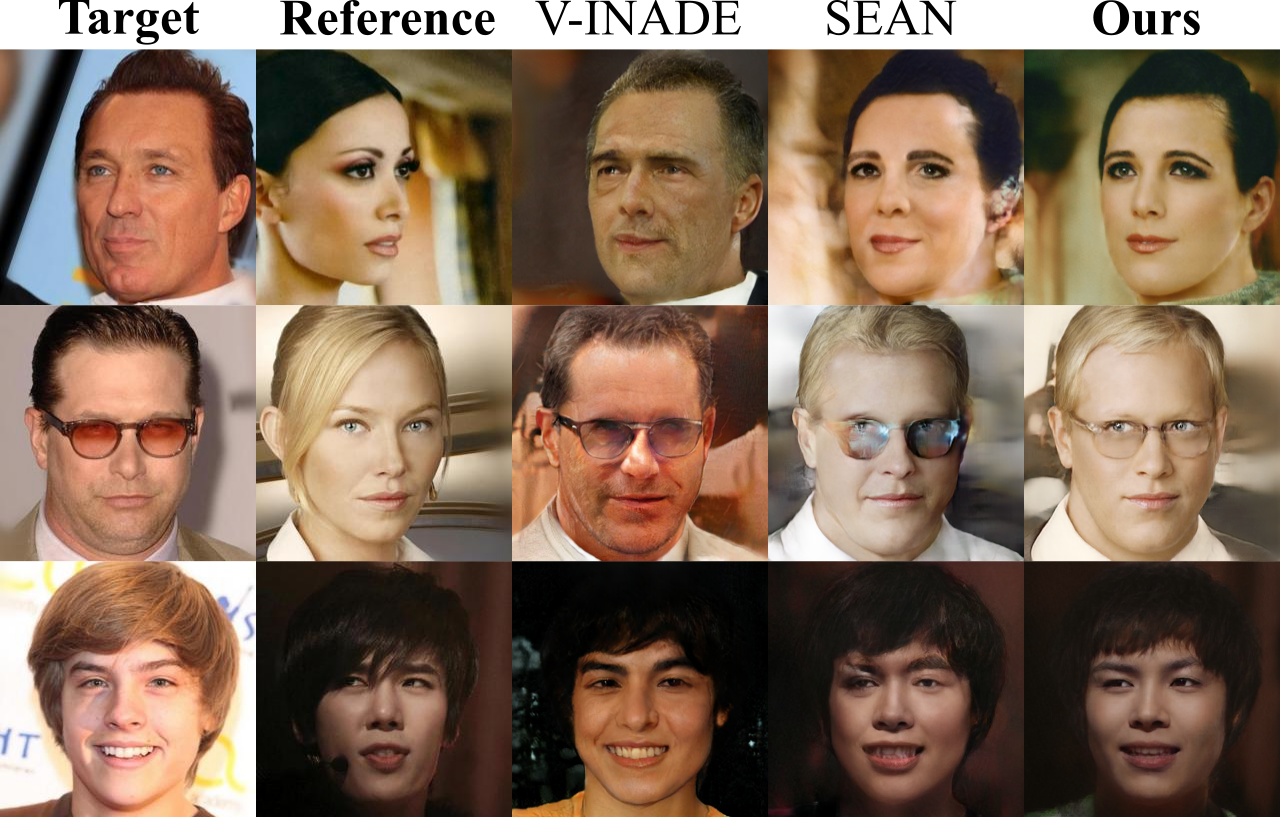}
    \caption{Comparison between V-INADE, SEAN and \methodacro{} when transferring the style of all face parts. Our method convincingly generates realistic results even if a specific style \textit{i.e.} opposite ear in the top row, eyeglasses in the middle row, and teeth in the bottom row, are absent in the reference image.}
    \label{fig:style}
\end{figure}

\subsection{Style transfer}
In Fig.~\ref{fig:style} we report some examples of style transfer of all face parts from one reference face to another one. Note that we did not put effort into picking particular examples, yet we chose some that best highlight the limitations of previous solutions. 
Our method achieves significantly more realistic transfer results even in very challenging cases. Our model is able to correctly integrate texture even when it is not present in the reference image but its semantic class exists in the target image (\textit{e.g.}~opposite ear in the top row, eyeglasses in the middle row, teeth in the bottom row). On the opposite, SEAN cannot handle such cases. This is possible thanks to the self- and cross-attention layers which learn shape-style correlations. 

\begin{figure}[t]
    \centering
    \includegraphics[width=0.98\linewidth]{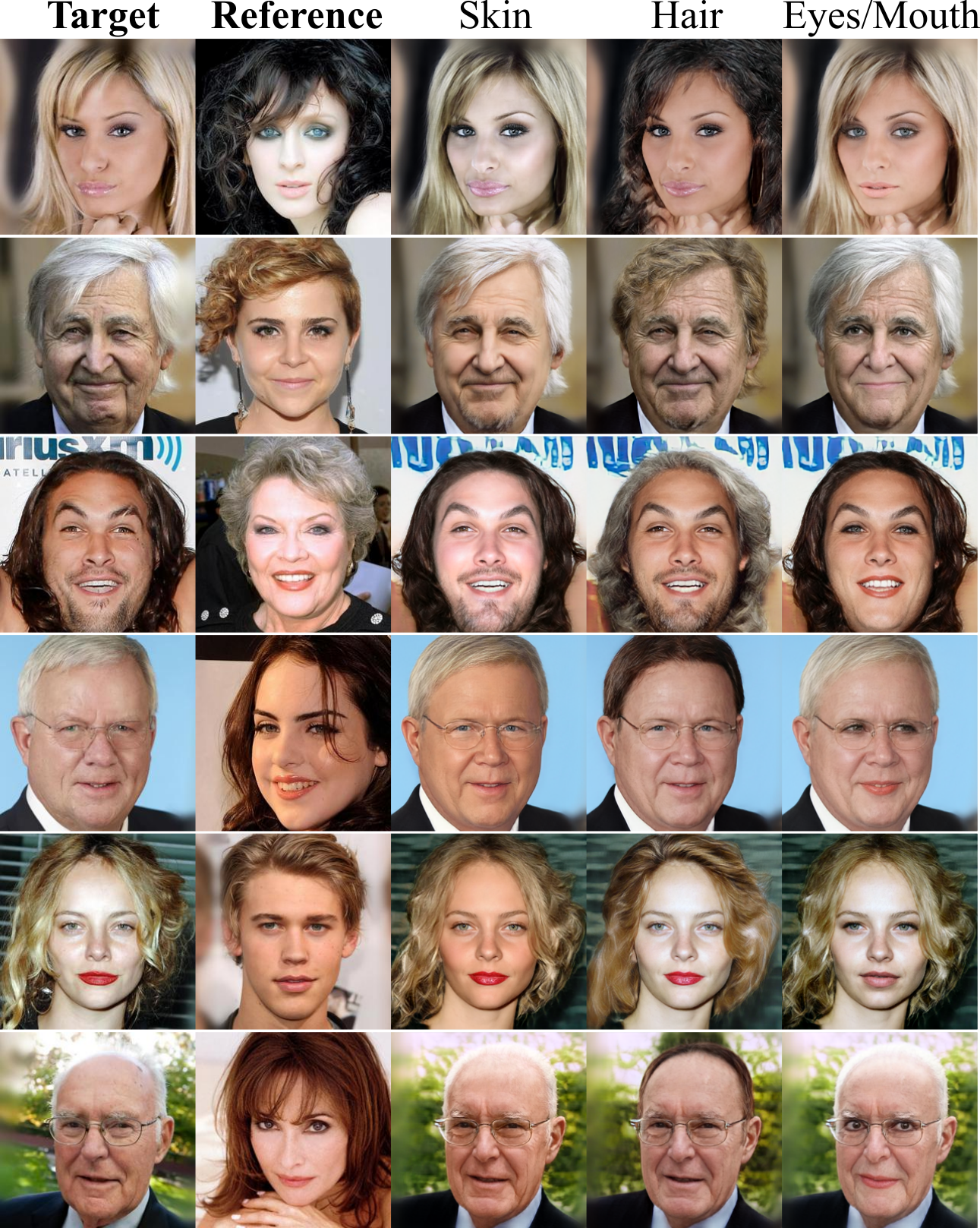}
    \caption{Style transfer of face parts. \methodacro{} can successfully perform local style manipulation even if no spatial information is used in the generator.}
    \label{fig:style_parts}
\end{figure}

\begin{figure}[t]
    \centering
    \includegraphics[width=0.98\linewidth]{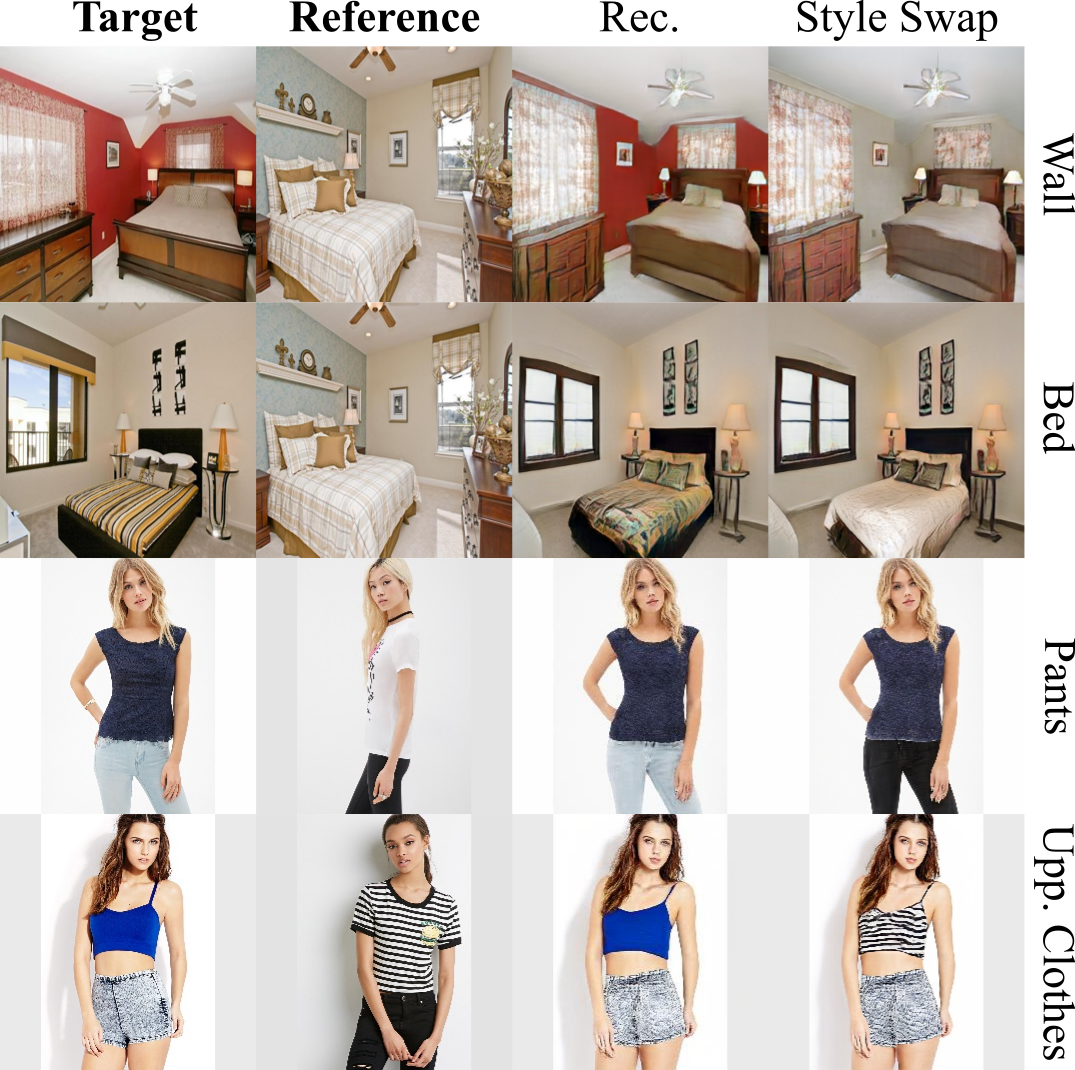}
    \caption{Style transfer of single semantic parts in ADE20k and DeepFashion.}
    \label{fig:ade_deepf}
\end{figure}

In Fig.~\ref{fig:style_parts} and~\ref{fig:ade_deepf} we also report some examples of style transfer at the class level. 
Our model can accurately apply local styles and generate realistic results even for challenging cases (\textit{e.g.} different poses) without sacrificing image consistency, and maintaining complex details such as global illumination coherence. 
We note this is a remarkable result given that we \textit{do not use any explicit spatial information} in the intermediate layers to guide the style mapping into the generator features.

\begin{figure}[t]
    \centering
    \includegraphics[width=.99\linewidth]{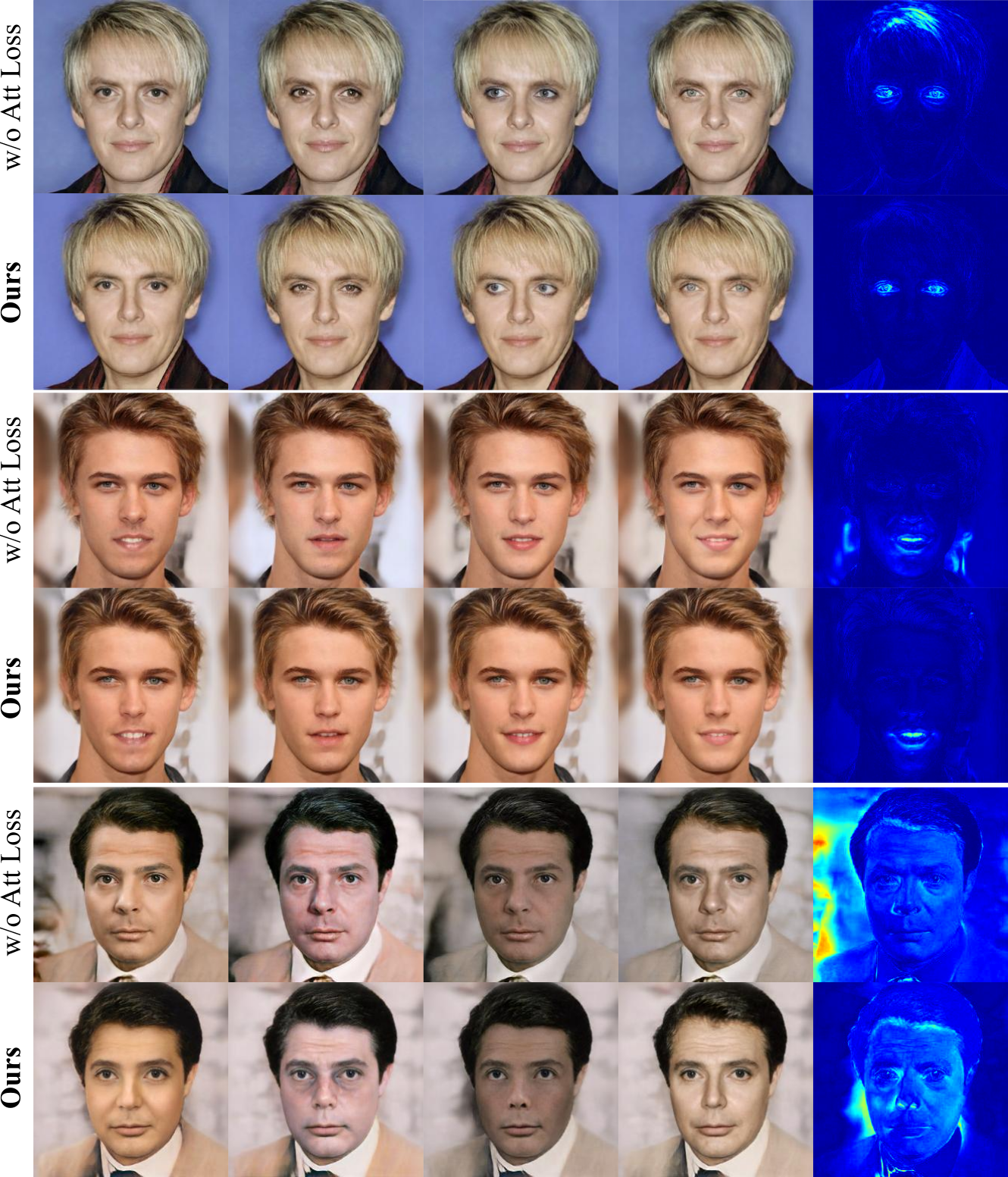}
    \caption{Examples of local style transfer with and without attention loss. The same target image is edited by transferring the style of a face part (top: eyes, middle: mouth, bottom: skin) from four different reference images. The heatmpas show the average $L1$ difference across the edited images and the original one. The attention loss leads to more localized editing.}
    \label{fig:swap_att}
\end{figure}

Finally, in Figure~\ref{fig:swap_att} we provide an example that makes the contribution of the attention loss explicit. We take one target image and apply the style of a face part taken from four different reference images. Then, we compute the average $L1$ difference and convert it to a heatmap. The latter clearly evidences that the attention loss leads to a more localized editing. Without it, the cross-attention tends to expand its influence beyond the part of interest. For example, in the top row, eyes are edited, but some manipulation to the hair is applied as well. The attention loss corrects this (second row).

\begin{figure}[!b]
    \centering    
    \includegraphics[width=\linewidth]{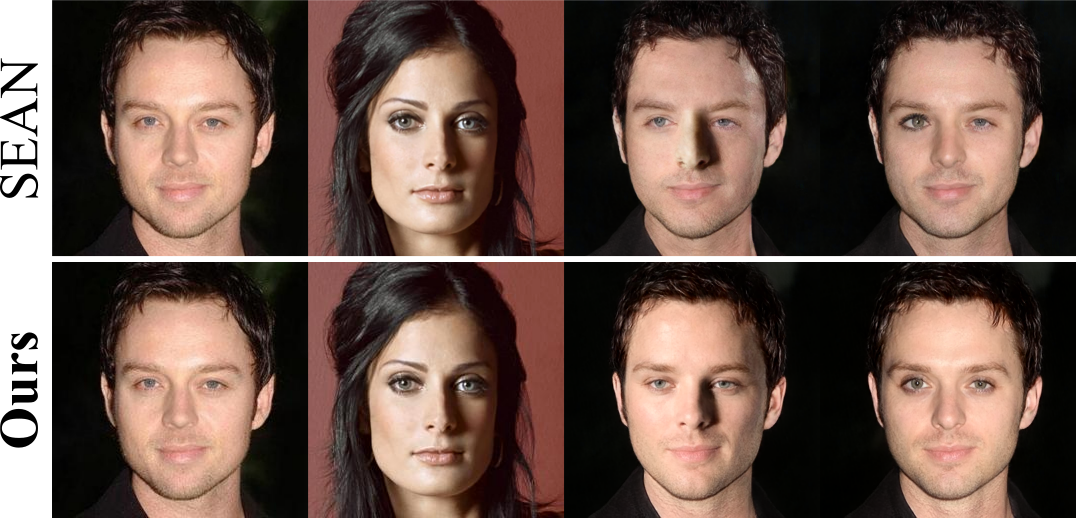}
    \caption{Transferring the style of local parts can lead to overall image inconsistencies, for example, if the target and reference image differ in illumination, or if changing highly correlated parts. Methods using spatially-adaptive normalization layers (i.e. SEAN) cannot handle such cases. Our solution fixes this effect to a large extent.}
    \label{fig:single_style_parts}
\end{figure}

\bigskip
\noindent\textbf{Fixing Image Consistency.}
Performing style transfer \--- or editing \--- of local features is a tricky task, as it can cause global inconsistencies to arise in the manipulated image. For example, this can happen when trying to mix the style of images with different illumination, or if trying to transfer the style of strongly correlated classes.
In Fig.~\ref{fig:single_style_parts} some examples are depicted where we compare the transfer of two parts, the nose and left eye. 
SEAN is able to precisely swap the style of the single parts yet at the cost of making the final image resulting overall unrealistic and inconsistent. Our method, instead, tends to fix this lack of consistency; in particular, it does so by, in one case (Fig.~\ref{fig:single_style_parts}, third column) removing the unnatural cast shadow by adjusting the overall illumination and shades, and, in the other case (Fig.~\ref{fig:single_style_parts}, fourth column) by adjusting the style of the symmetric eye, as eyes color is normally the same in human faces.
This behavior is two-faced: on the one hand, it restricts the ability of the model to precisely apply local styles; on the other, it maintains a high level of overall visual quality and realism. Note that all the results in Fig. ~\ref{fig:single_style_parts} are obtained using attention loss, which retains a good trade-off between generation quality and controllability.

\subsection{Shape transfer and manipulation}
\begin{figure}[t]
    \centering
    \includegraphics[width=\linewidth]{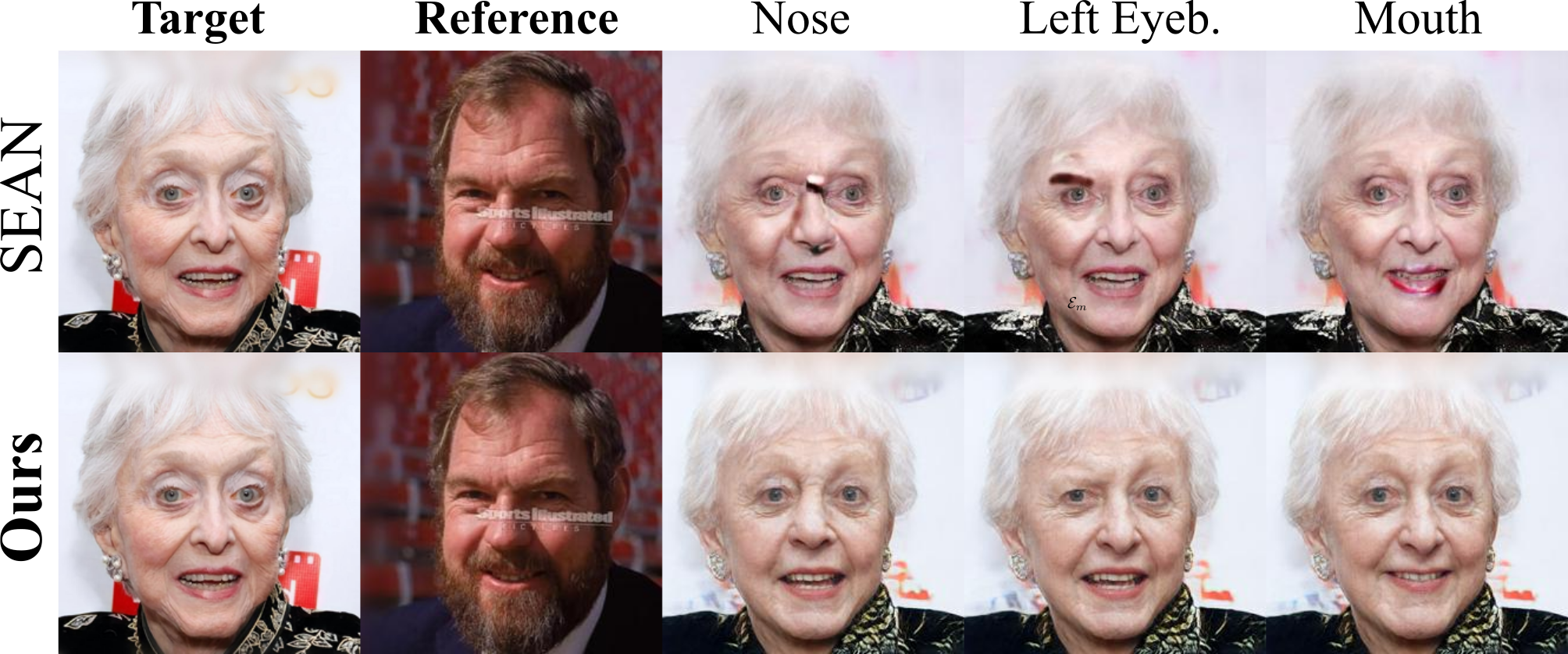}
    \caption{Shape transfer comparison between SEAN and our final architecture. SEAN cannot handle automatic shape changes, and require manual intervention to fix the mask. Our method equipped with the mask embedder $\mathcal{E}_m$ can instead automatically do so.}
    \label{fig:swap_parts}
\end{figure}

Unlike previous methods, in our framework the semantic mask is embedded into a set of class-wise latent codes. We briefly anticipated that in certain circumstances, other than resulting in improved reconstruction, this allows for additional advantages such as the possibility of globally or even locally manipulating the shape. In Fig.~\ref{fig:swap_parts} we report some qualitative examples of \textit{shape transfer}, \textit{i.e.} changing the embedding of a semantic class from a reference mask to a target one. Again, we compare our solution with SEAN as representative of previous SPADE-based methods. 
Using the raw mask induces several artifacts (holes) in the generated images even for slight misalignment of face parts. This spatial inconsistency cannot be handled by methods that use SPADE to explicitly inject the style into specific spatial locations. 
On the other hand, our solution allows for fixing such minor inconsistencies, opening the way to fully automatic control of both style and shape. Quantitatively, this advantage is confirmed by comparing the columns FID and FID-sh in Table~\ref{tab:shape_transfer}, which reports the FID score obtained on reconstructed images versus shape-manipulated ones. 


\begin{figure}[!h]
    \centering
    \includegraphics[width=\linewidth]{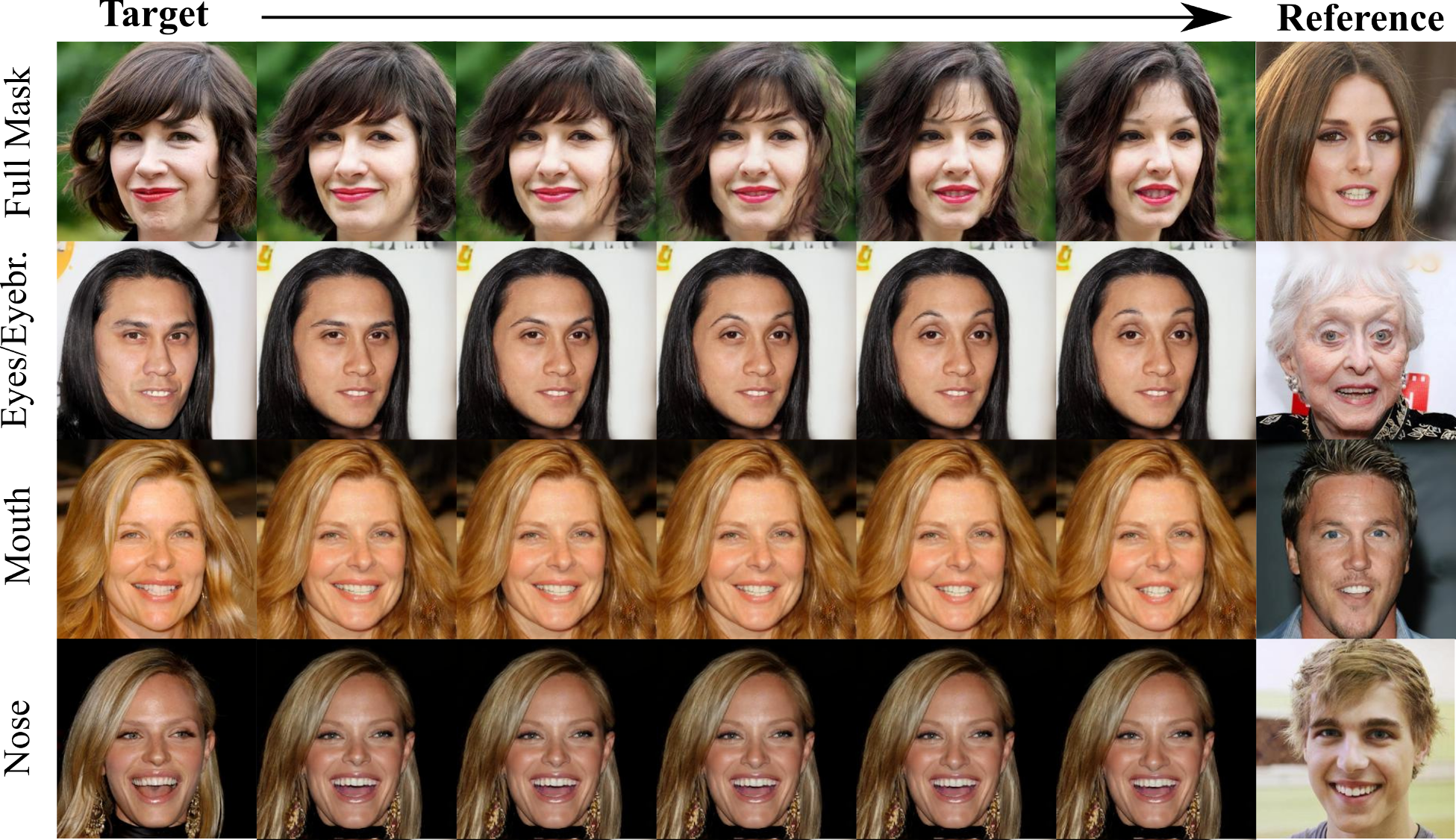}
    \caption{Interpolation between two mask embeddings. The generated images naturally transition between two shapes.}
    \label{fig:interpolation}
\end{figure}

\bigskip
\noindent\textbf{Shape interpolation}
Other than transferring the shape of face parts from one image to another, we can extend this ability to perform geometry \textit{interpolation}. The only other method able to do so is MaskGAN~\cite{lee2020maskgan}. However, it can only perform global mask interpolations and needs an additional alpha-blender to refine the generated images.  Differently, we are able to linearly interpolate any arbitrary channel $j$ from two masks $\mathcal{M}^1$ and $\mathcal{M}^2$, and generate an interpolated mask embedding as $m^{int}_j = \alpha\cdot\mathcal{E}_m(\mathcal{M}^1_j) + (1-\alpha)\cdot\mathcal{E}_m(\mathcal{M}^2_j)$. Fig.~\ref{fig:interpolation} reports some examples of face images generated in this way. 
Interestingly, when interpolating the whole mask, the 3D head pose is also changed. On the one hand, this is an intriguing effect that could open the way to novel applications; on the other, it can be problematic when manipulating local parts, in which case inconsistencies still occur if not properly handled. 

\subsection{Comparison with StyleGAN-based methods}\label{subsec:stylegan}
In the field of human face generation, StyleGAN~\cite{karras2019style} and all its subsequent improvements are among the most effective unconditional face generation models to date. Indeed, almost all recent methods for face generation and manipulation are based on a pre-trained StyleGAN model. Being purely generative though, they cannot be directly used for reconstructing and editing \textit{real faces}. However, a workaround has been found, which technique is referred to as \textit{GAN-Inversion}~\cite{zhu2020domain}. It is based on the idea of applying some optimization for finding the embedding in the StyleGAN latent space that best approximates a given real face image. Once this latent is found, it is possible to edit the reconstructed face by applying latent manipulation techniques.

Using GAN-Inversion, remarkable results have been achieved. Yet, a question is whether the inversion process is sufficiently accurate to \textit{(i)} produce an image where the identity of the subject is sufficiently preserved, \textit{(ii)} maintain the level of quality and realism achieved in a purely generative setting and \textit{(iii)} be applied to out-of-domain data \textit{i.e.} faces not included in CelebA-HQ. In this section, we aim at comparing the results of our method against the most recent alternative based on StyleGAN, which is SemanticStyleGAN~\cite{shi2021SemanticStyleGAN}. This specific approach also performs local face editing based on a semantic mask. Results are reported in Table~\ref{tab:stylegan}. The performance of SemanticStyleGAN drops dramatically when applied in a GAN-inversion setting. The quality of the generated images decreases significantly as evidenced by the FID score which increases to 26.83 from 6.42 (value taken from~\cite{shi2021SemanticStyleGAN}) of the fully generative setting. Also, the ability of retaining the identity of the subject is compromised. We measure the latter using the Face Recognition Similarity (FRS) score: it is computed by embedding both real and reconstructed (inverted) images using an InceptionV3 pre-trained on VGG-Face2 dataset~\cite{cao2018vggface2}, and then computing the cosine similarity between embedding pairs. Our \methodacro{} scores a way higher similarity, indicating the identity is more effectively preserved. This is an essential feature for methods that aim at accurate editing of real images.

\begin{figure}
    \centering
    \includegraphics[width=\linewidth]{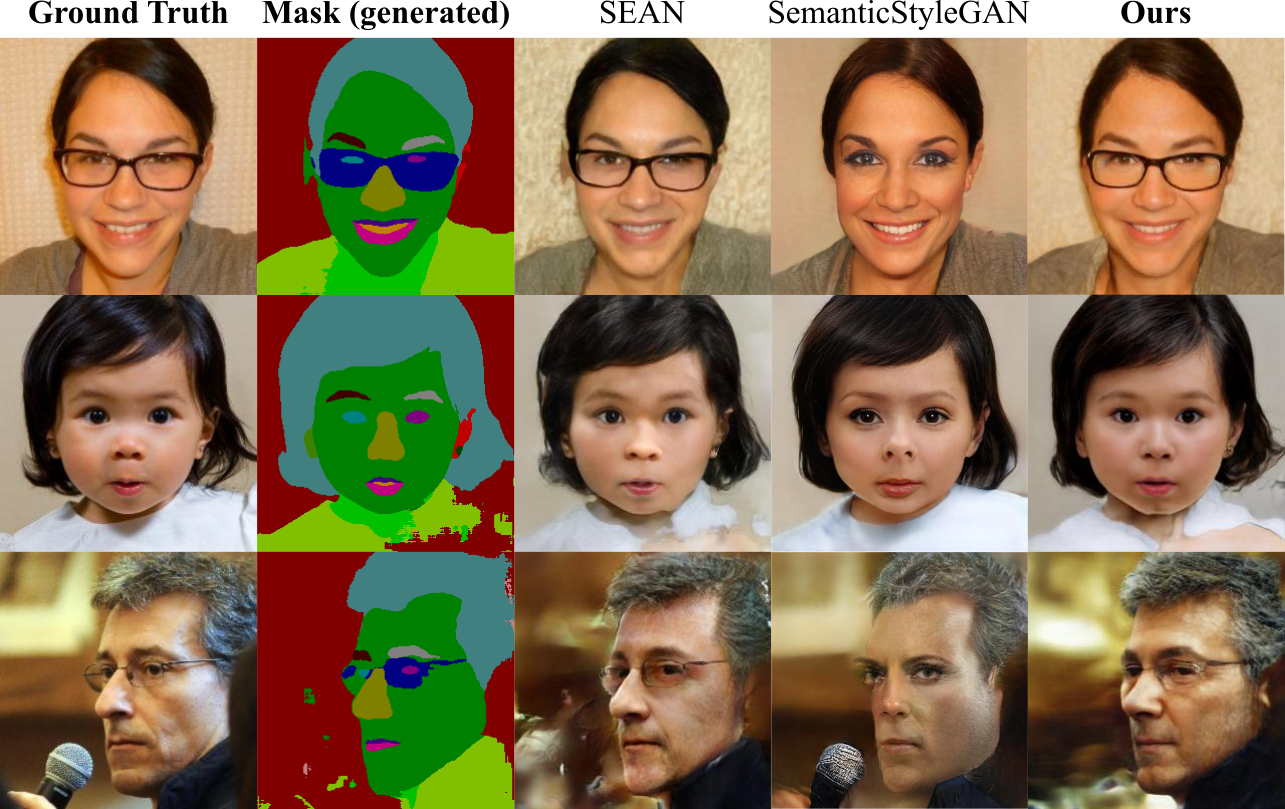}
    \caption{Results obtained on FFHQ with a model trained on CelebA-HQ.}
    \label{fig:itw}
\end{figure}

Methods based on StyleGAN are extremely promising in the generation task, yet these results evidence that they still fall short when manipulating a real image is the goal as they fail to apply the inversion process in a sufficiently accurate way. Another drawback is related to the worse generalization to out-of-distribution data. In Figure~\ref{fig:itw}, we qualitatively compare \methodacro{}, SemanticStyleGAN and SEAN on some images from FFHQ~\cite{karras2019style}, a commonly used dataset for training generative models. Semantic masks are not provided in FFHQ, so we employed the same face parser of Section~\ref{sec:experiments} to generate them. All the three models were trained on CelebA-HQ. SemanticStyleGAN fails dramatically in maintaining the identity traits of the subjects, mostly for severely under-represented classes such as the eyeglasses (top row), babies (middle row) or non-frontal faces (bottom row). SEAN instead can better preserve the identity and handle unseen samples, but results less robust to noise in the semantic masks. Our model more robustly handles detection noise as compared to SEAN. This is another advantageous result of embedding the mask into a latent vector and relying on attention mechanism.

\begin{table}[]
    \centering
    \begin{tabular}{c|c|c}
    \toprule
       \textbf{Method}  & \textbf{FID} $\downarrow$ & \textbf{FRS} $\uparrow$ \\
         \midrule
       Semantic StyleGAN~\cite{shi2021SemanticStyleGAN} & 26.83 & 0.68 \\
       \hline
       Ours & \textbf{15.82} & \textbf{0.81} \\
       \bottomrule
    \end{tabular}
    \caption{Comparison against Semantic StyleGAN on CelebMask-HQ.}
    \label{tab:stylegan}
\end{table}

\subsection{Subjective Evaluation}
\begin{figure}[!b]
    \centering
    \includegraphics[width=0.8\linewidth]{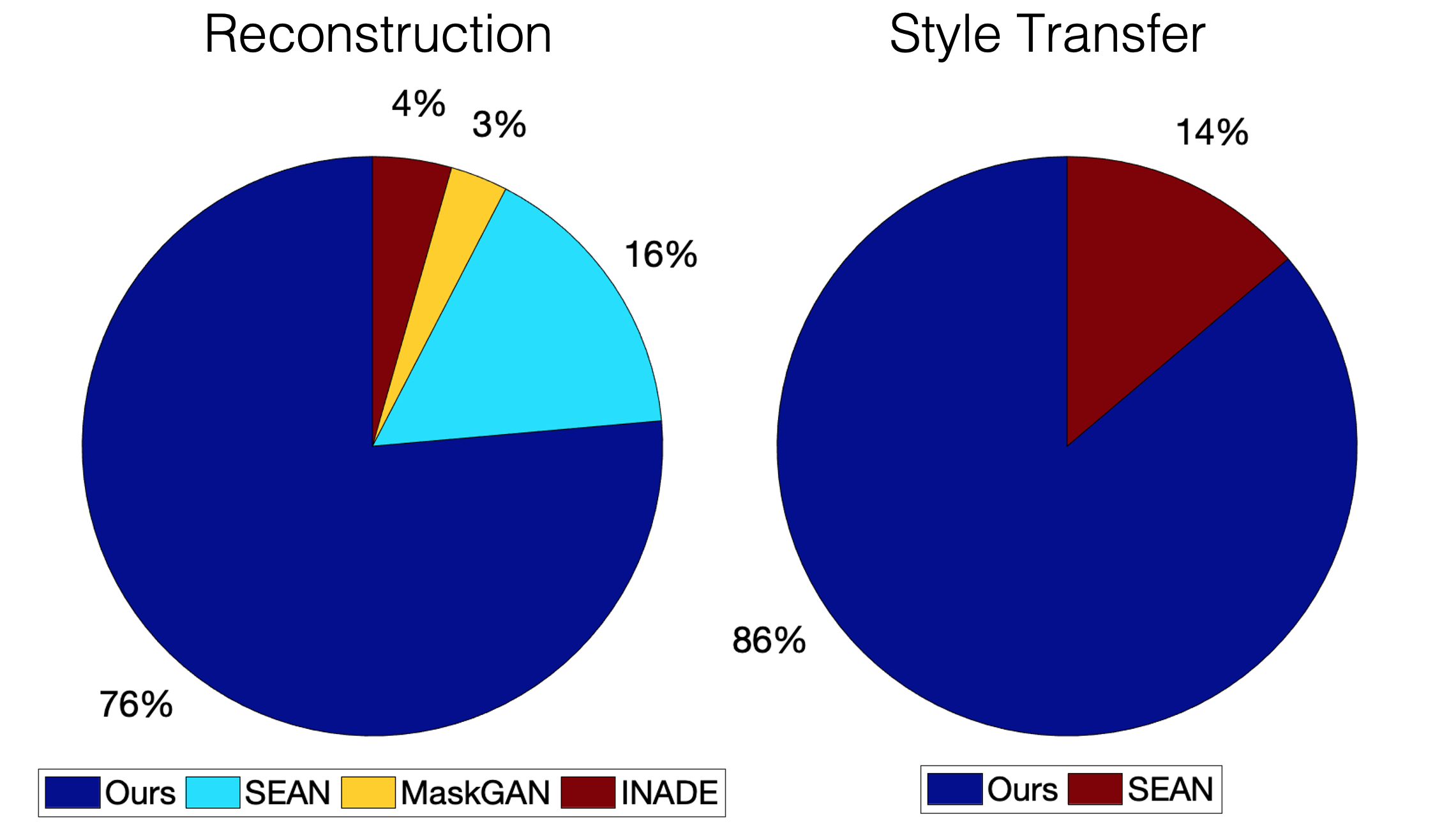}
    \caption{Subjective evaluation. Both charts show \methodacro{} is preferred by users, who selected our results as the most realistic in the majority of cases.}
    \label{fig:subj_eval}
\end{figure}

In this section, we report some results of a subjective evaluation conducted on CelebMask-HQ.
Following the protocol of~\cite{tan2021diverse} or~\cite{zhu2020semantically} (20 images are shown to 20 participants), we recruited 30 volunteers, both experts and non-experts in deep generative models, and asked them to select the most realistic result among those generated by different methods given the same input on 30 randomly selected images. The average percentage of times that users selected a specific method is reported in Fig.~\ref{fig:subj_eval}~(Reconstruction). Participants picked our results as the most realistic in the majority of cases. Using attention mechanisms to condition the image generation has the advantage that global style correlations are preserved more faithfully than using adaptive-normalization layers, which treat each class (or instance) independently, eventually making the generated image perceived as less natural.

In addition, we also asked the participants to evaluate the results for the task of style transfer. 
Here we compared only against SEAN. V-INADE was excluded; even if equipped with the variational style encoder, its goal is to approximate the class distribution so as to generate diverse outputs. 
As shown also in Fig.~\ref{fig:style}, it does not perform well on style transfer as it was not designed for that goal.
We applied the style of a reference image $A$ to the semantic mask of another image $B$, for 10 random image-mask pairs. Results reported in Fig.~\ref{fig:subj_eval}~(Style Transfer) set off our method as the most realistic.  

\subsection{Ablation study}\label{subsec:ablation}
Finally, in Table~\ref{tab:ablation} and Table~\ref{tab:shape_transfer}, we report a detailed ablation study. We performed three sets of experiments, exploring several configurations for the style encoder and the generator, respectively, and for the task of shape transfer.

\paragraph{Style Encoder}
We first trained a version of our architecture equipped with the style encoder of SEAN~\cite{zhu2020sean} (Ours - SEAN $\mathcal{E}_s$). Compared to pure SEAN (FID 18.72), our FID score is lower (FID 17.81), indicating the cross-attention-based generator leads to better results given the same style features. Then, we tested a version of our multi-scale style encoder without group convolutions (Ours w/o GC). Compared to our full model, gathering features with group convolutions enables capturing class-specific texture details at different levels of granularity, leading to more realistic generated images.

\begin{table}[]
    \centering
    \begin{adjustbox}{width=0.8\columnwidth,center}
    \begin{tabular}{@{}c|c}
    \toprule 
    \textbf{Style Encoder} & \textbf{FID} $\downarrow$ \\  
    \hline
    Ours - SEAN $\mathcal{E}_s$ & \underline{17.81} \\
    \hline
    Ours w/o GC & 16.42 \\
    \hline
    \hline
    \methodacro{} & \textbf{15.84} \\
    \bottomrule  
    \end{tabular}
    \quad
    \begin{tabular}{@{}c|c}
    \toprule
    \textbf{Generator} & \textbf{FID} $\downarrow$ \\
    \hline
    Ours w/o $\mathcal{E}_m$ & 18.65 \\
    \hline
    Ours w/o CA & 18.14 \\
    \hline
    \hline
    \methodacro{} & \textbf{15.84} \\
    \bottomrule   
    \end{tabular}

    \end{adjustbox}
    \caption{Ablation study on CelebMask-HQ. ``\emph{CA}'' stands for cross-attention. ``\emph{GC}'' stands for group convolutions.}
    \label{tab:ablation}
\end{table}

\paragraph{Generator}
Regarding the generator, we observe that removing the mask embedder (Ours w/o $\mathcal{E}_m$) worsens the quality. Indeed, embedding the mask to remove spatial information demonstrated beneficial to better capture shape-style correlations with cross-attention modules. Finally, we replaced cross-attention layers with SPADE layers (Ours w/o CA). These could not exploit well the multi-resolution style codes as compared to cross-attention. 

\paragraph{Shape transfer}
In Table~\ref{tab:shape_transfer}, we report a quantitative comparison for the task of shape transfer. Those are computed by swapping a random part (random seed is fixed for all methods) across pairs of images, without imposing any constraints for choosing the pairs. Results are crystal clear; the FID in case of shape transfer (FID-Sh) of both SEAN and V-INADE increases significantly, while it remains stable for our method, supporting the qualitative results in Fig.~\ref{fig:swap_parts}.

\begin{table}[!h]
    \centering
    \begin{adjustbox}{width=0.5\columnwidth,center}
    \begin{tabular}{@{}c|c|c@{}}
    \toprule
    \textbf{Method} & \textbf{FID} $\downarrow$ & \textbf{FID-Sh} $\downarrow$ \\
    \hline
      SEAN & 18.72  & 20.04 \\
      \hline
      V-INADE & 17.49  & 31.16 \\
      \hline
      \hline
      \methodacro{} & \textbf{15.84} & \textbf{15.90} \\
      \bottomrule
    \end{tabular}

    \end{adjustbox}
    \caption{Results for shape (Sh) transfer on CelebMask-HQ.}
    \label{tab:shape_transfer}
\end{table}

\smallskip
The above experiments highlight how the proposed model is a promising alternative with respect to SPADE and variants as it allows for \textit{(i)} more versatility in designing proper style features, \textit{(ii)} embedding the semantic mask into latent vectors, leading to more freedom in geometry manipulation and \textit{(iii)} improved image quality without sacrificing editing control.  

\subsection{Diversity-driven generation}\label{subsec:diversity}
The main goal of this work was to explore a new, alternative architecture for semantic image synthesis guided by a reference style image. However, the proposed generator architecture based on spatial transformer layers in place of SPADE-like normalization is general in its purpose and can be potentially used also in a fully-generative setting. Clearly, the latter requires specific training strategies to force both quality and diversity in the generated images, which is beyond the scope of this work. Nevertheless, in this section we report the results of an additional set of experiments aimed at showing that we can train \methodacro{} in the diversity-driven setting and obtain results that are on-par with respect to state-of-the-art methods specifically tailored for such task.

\begin{figure}
    \centering
    \includegraphics[width=\linewidth]{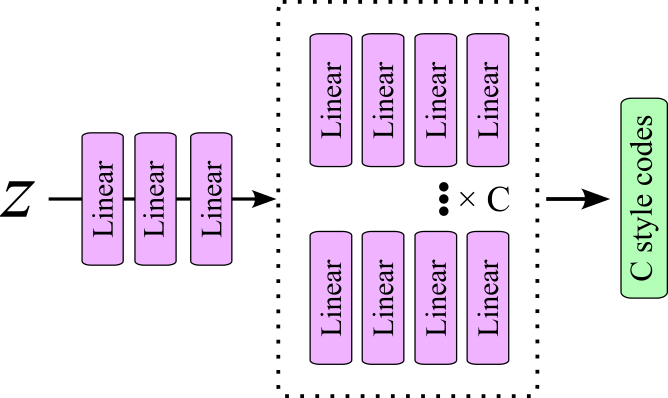}
    \caption{Mapping network $\mathcal{M}_s$ architecture: a noise vector $z \sim \mathcal{N}(0,\,1)$ is processed to obtain a set of $C$ style codes.  }
    \label{fig:mapping_net}
\end{figure}

To achieve this, we exploit the pre-trained \methodacro{} architecture as described in Section~\ref{sec:architecture}, but we substitute the style encoder $\mathcal{E}_s$ with a mapping network $\mathcal{M}_s$ designed to map a noise vector into the latent style codes (see Fig. \ref{fig:mapping_net}). The design of the mapping network $\mathcal{M}_s$ is quite simple: it takes a noise vector $z \sim \mathcal{N}(0,\,1)$ as input, which passes through 3 linear layers, each outputting a latent vector of size $512$. Then, the output of the third layer branches into $C$ paths, where $C$ is the number of semantic classes. Each branch is composed of 4 linear layers. The first three output a latent vector of size $512$, while the last one's output has size $1,280$, which is needed to match the size of the multi-scale style codes resulting from $\mathcal{E}_s$ \textit{i.e.} $256 \times 5$. The output of the mapping network $\mathcal{M}_s$ becomes the $Key$ and $Value$ for the cross-attention layers; recalling Eq.~\eqref{eq:attention}, that is:
\begin{equation}
    Q=W_Q^{(i)} \cdot \phi^{(i)}, 
    K = W_K^{(i)} \cdot \mathcal{M}_s(z), 
    V = W_V^{(i)} \cdot \mathcal{M}_s(z)
\end{equation}

To train the mapping network, we exclude the style encoder $\mathcal{E}_s$, keep the generator $G$ frozen and re-initialize the discriminator. The training is guided by only two losses: the discriminator loss $\mathcal{L}_D$ and a diversity loss $\mathcal{L}_{dv}$. The latter maximizes the $L1$ discrepancy between two images $I_1, I_2$ generated from two noise vectors $z_1,z_2$ given the same semantic mask $m$, that is $\mathcal{L}_{dv} = - \Vert G(\mathcal{M}_s(z_1),m) - G(\mathcal{M}_s(z_2),m)\Vert _1$. The network is trained for 100 epochs, using the same parameters as described in Section~\ref{sec:experiments}.

\begin{figure}[!t]
    \centering
    \includegraphics[width=\linewidth]{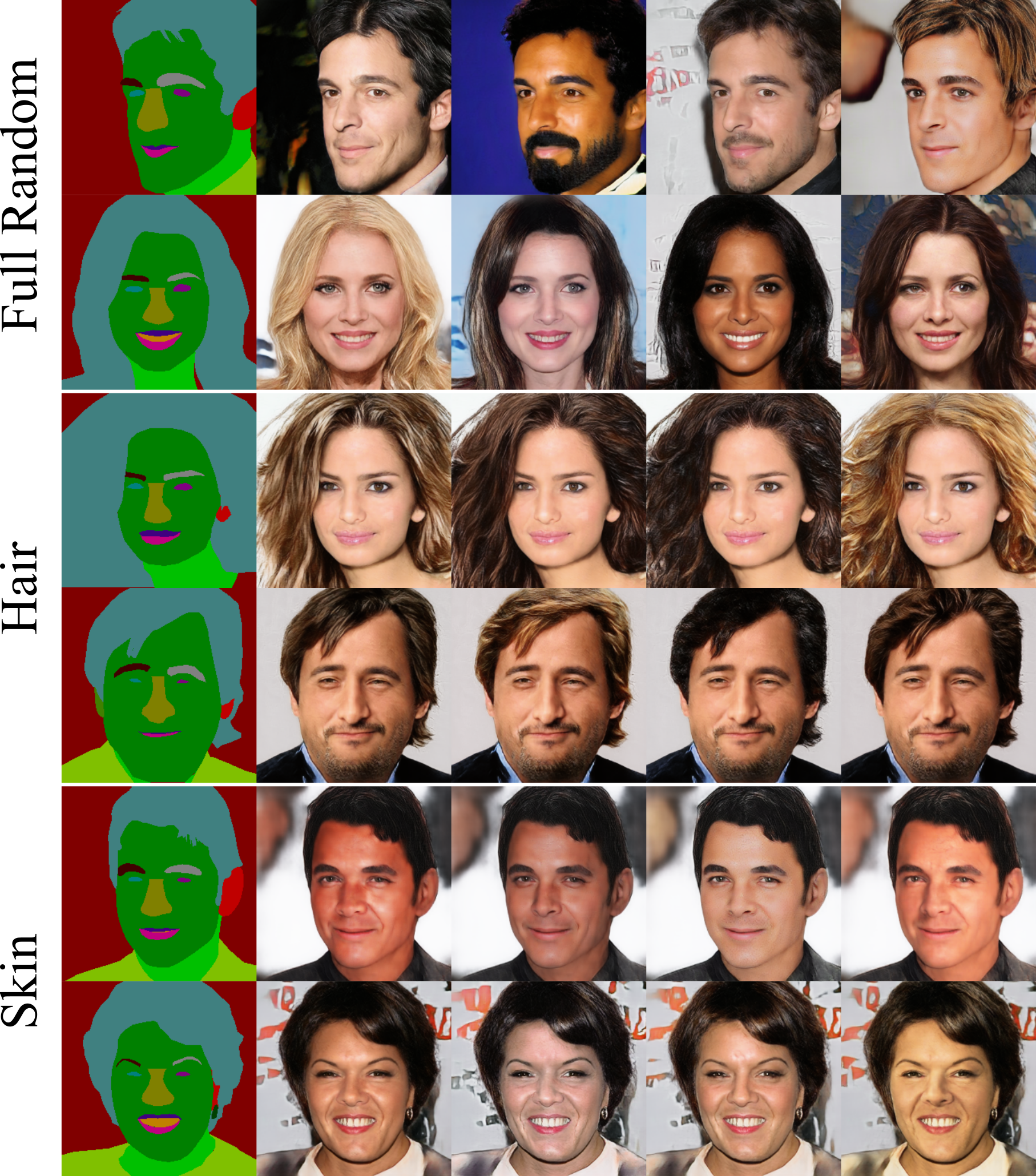}
    \caption{Images generated with our model trained for diversity. In the first two rows all the 19 style codes corresponding to the CelebMask-HQ semantic classes are generated by the mapping network $\mathcal{M}_s$. In the remaining rows, we kept all the style codes fixed except hair and skin to demonstrate the disentanglement capability of the model.}
    \label{fig:diversity}
\end{figure}

\begin{table}[!t]
    \centering
    \begin{tabular}{c|c|c}
    \toprule
    \textbf{Method} & \textbf{FID} $\downarrow$ & \textbf{LPIPS} $\uparrow$ \\
    \hline
    SC-GAN~\cite{wang2021image} & 20.85 & 0.170  \\
    INADE~\cite{tan2021diverse} & \textbf{18.31} &  0.365 \\
    SemDiffusion~\cite{wang2022semantic} & \underline{19.22} & \textbf{0.422} \\
    \hline    
    \textbf{Ours} & 19.79 & \underline{0.367} \\
    \bottomrule   
    \end{tabular}
    \caption{Comparison with the state-of-the-art in terms of FID, and LPIPS. Best results in bold, second best underlined.}
    \label{tab:celeba-rec-div}
\end{table}

Results on CelebMask-HQ are shown in Table~\ref{tab:celeba-rec-div}, in comparison with three state-of-the-art approaches, namely INADE~\cite{tan2021efficient}, SC-GAN~\cite{wang2021image} and SemanticDiffusion~\cite{wang2022semantic}. Note that, differently from the methods in Table~\ref{tab:celeba-rec}, these are purely diversity-driven and do not use any reference images. In terms of image quality (FID), our method's performance are comparable to both previous GANs equipped with SPADE layers (INADE), other custom modules such as conditional convolutions (SC-GAN) and SemanticDiffusion, which also employs SPADE to condition the diffusion process with the semantic mask. In terms of diversity (LPIPS~\cite{zhang2018unreasonable}) instead, SemanticDiffusion stands out. About the latter, metrics were computed on images generated with $1,000$ diffusion steps to ensure the best possible quality. However, generating the whole test set (2K images) in this way, took approximately \textit{11 days} on a NVIDIA A100 GPU (8 minutes per image). We also tried using the DDIM~\cite{song2020denoising} technique to reduce the diffusion steps. Using only 50 steps, generating a single image takes approximately 10 seconds, yet the FID grows from $19.22$ to $32.96$. Our \methodacro{} instead takes $0.07$ seconds to generate one image. Whereas using a diffusion models provided excellent results, the computational burden makes its usage quite unpractical. Moreover, our solution allows for a nice feature not shown by previous methods, that is generating diverse and disentangled styles for different semantic classes. For example, one can keep the style of some parts fixed while only generating that of a specific class, such as skin or hair, allowing for a detailed generation control over local classes. Some qualitative results are shown in Figure~\ref{fig:diversity}.

\section{Limitations and Conclusions}\label{sec:limitations}
In this section, we discuss some issues, limitations, and peculiar behaviors of our solutions. 
First, although shape transfer results clearly show that our model can manipulate the shape significantly better than previous works, in case of strong misalignment or large translation, gaps and inconsistencies still occur. This prevented us from successfully applying shape transfer on other datasets (CelebMask-HQ is particularly suitable as face images are inherently quite well aligned).
Nevertheless, the ability of our model to learn to apply shape changes without specific supervision, achieved thanks to the proposed paradigm change, is valuable and can open the way to further improvements. 
From a different perspective, as can be noted from Fig.~\ref{fig:style}, using cross-attention layers in place of adaptive normalization ones is advantageous as they can fix the generation process for an overall increased image consistency. 
On the one hand, in cases where parts styles are strongly correlated, cross-attention can prevent full-style control over single classes (eyes color in Fig.~\ref{fig:single_style_parts}). 
On the other hand, style transfer outcomes look significantly more realistic and consistent.

In conclusion, in this paper we proposed a novel architecture that uses attention mechanisms as an alternative to SPADE-like normalization layers to perform semantic image synthesis. In addition, we employ an attention loss to force the attention in the model to match the semantic mask. This has the effect of reducing entanglement between different styles and better maintaining the semantic mask shape in the generated samples. We provided a detailed analysis of its values and limitations with respect to prior art, and a preliminary solution to the new task of automatic manipulation of local geometry. 

\section{Acknowledgment}
This work was partially supported by PRIN 2020 “LEGO.AI: LEarning the Geometry of knOwledge in AI systems”, grant no. 2020TA3K9N funded by the Italian MIUR. We also acknowledge the CINECA award under the ISCRA initiative (project: IsCa5\_X-SYS), for the availability of high performance computing resources and thank Andrea Pilzer and Giuseppe Fiameni of NVIDIA for the support.

\bibliographystyle{IEEEtran}
\bibliography{bibliography}

\end{document}